# The Legislative Recipe: Syntax for Machine-Readable Legislation

*Megan Ma, Bryan Wilson*

Legal interpretation is a linguistic venture. In judicial opinions, for example, courts are often asked to interpret the text of statutes and legislation. As time has shown, this is not always as easy as it sounds. Matters can hinge on vague or inconsistent language and, under the surface, human biases can impact the decision-making of judges.[1] This raises an important question: what if there was a method of extracting the meaning of statutes consistently? That is, what if it were possible to use machines to encode legislation in a mathematically precise form that would permit clearer responses to legal questions? This is the fundamental basis of the Rules as Code initiative.

To recall, Layman E. Allen lamented about ambiguity in legal drafting owed to syntactic uncertainties.[2] In his fascinating study, he deconstructs an American patent statute and notices immediately the complexity with the word 'unless.' He asks whether the inclusion of 'unless' asserts a unidirectional or a bidirectional condition.[3] That is, does the clause mean (a) if not x then y; or (b) if not x then y **and** if x then not y?

Though nuanced, Allen exposes an ambiguity that muddies the legal force of the statute. An interpretation of 'unless' as a bidirectional condition raises the question of what 'not y' would mean. In this particular case, this could affect whether exceptions are possible in determining patent eligibility. In short, for Allen, legislative language must have a clear structure.

This article attempts to unpack the notion of machine-readability, providing an overview of both its historical and recent developments. The paper will reflect on logic syntax and symbolic language to assess the capacity and limits of representing legal knowledge. In doing so, the paper seeks to move beyond existing literature to discuss the implications of various approaches to machine-readable legislation. Importantly, this study hopes to highlight the challenges encountered in this burgeoning ecosystem of machine-readable legislation against existing human-readable counterparts.

---

[1] In "Emotional Judges and Unlucky Juveniles" Ozken Eren and Naci Mocan found that upset losses (i.e., losses by LSU football team when it was expected to win) increase the sentence length imposed by judges on juvenile defendants. American Economic Journal: Applied Economics Vol. 10, No. 3, (2018). https://www.aeaweb.org/articles?id=10.1257/app.20160390

[2] Layman E. Allen, "Language, Law, and Logic: Plain Legal Drafting for the Electronic Age," B. Niblett (ed.) *Computer Science and Law* 76 (1980).

[3] *Id*. at 77.

### A. Historical Roots: Symbolic Logic

The code of Hammurabi[4] is frequently used as an example of how the law has changed in form in order to improve access to the legal system, lead to more predictable legal outcomes, and to promote transparency. Through the adoption of form, law can be understood as a body of knowledge that over time has come to inform behavior through the production, dissemination, and evaluation of the rules. Lawrence Lessig and Alex "Sandy" Pentland each have highlighted this with the notions that code is law, and law is an algorithm. More recently, concrete efforts have highlighted the business, technical, and legal interest in combining law and code to do things that each on their own is unable to do, including: analyzing public commit patterns of openly maintained cryptocurrencies[5] and demonstrating how code could function as an antitrust mechanism to supplement areas where the rule of law does not fully apply.[6] As a sort of synthetic or artificial intelligence of its own, the purpose of law has always been to measurably improve outcomes of its subjects. Whether aided by machines or undertaken entirely by humans (a technology of our own), this does not happen without the socialization of linguistic models for rules.[7]

These ideas are not new. The ancestry dates back to twelfth century logicians reflecting on the use of mathematically precise forms of writing. In the mid-1930s, German philosopher, Rudolf Carnap, reflected on a logical syntax for language.[8] His argument is that logic may be revealed through the syntactic structure of sentences. He suggests that the imperfections of natural language point instead to an artificially constructed symbolic language to enable increased precision. Simply put, it is treating language as a calculus.[9]

---

[4] Michael Genesereth, "The Legacy of Hammurabi" (Mar. 17, 2021), available at: https://law.stanford.edu/2021/03/17/the-legacy-of-hammurabi/.

[5] Lorenzo Lucchini et. al, "From code to market: Network of developers and correlated returns of cryptocurrencies" *Science Advances* (2020) available at: https://advances.sciencemag.org/content/6/51/eabd2204.abstract?fbclid=IwAR0REZe0VCRCZwySwR6C_FcG6FB 6DZQLAFwhGyF6QTvbZmoOrMGxAONpk4k.

[6] Thibault Schrepel and Vitalik Buterin, "Blockchain Code as Antitrust" *Berkeley Technology Law Journal* (2020) https://papers.ssrn.com/sol3/papers.cfm?abstract_id=3597399.

[7] M.J. Sergot et. al, *The British Nationality Act as a Logic Program*, 29 Communications of the ACM 370-386 (1986) available at: http://www.doc.ic.ac.uk/~rak/papers/British%20Nationality%20Act.pdf

[8] Rudolf Carnap, *Logical Syntax of Language* 2 (Routledge English ed. reprint, 2014).

In this perspective, there is no consideration of language for the intentions of meaning and interpretation. Merely, logical syntax is concerned with structure and is void of content.[10] Though Carnap concedes that syntax belongs to the scientific study of language that enables mathematical calculation, this approach must be distinguished from semantics, or semasiology. For Carnap, syntax importantly builds a system of reference. In an analogy with the "complicated configurations of mountain chains, rivers, frontiers, and the like," geographical coordinates are mathematical constructions that act as informative measurements of comparison to reveal and analyze the behaviors of its 'natural' existence.[11] Symbolic language, therefore, acts to investigate and identify consistencies and contradictions in language for the purpose of clarifying its logical properties.

Since the 1950s, Allen had argued for the inclusion of symbolic logic to develop a systematic method of drafting. The transformation of an ordinary statement to a "systematically pulverized form"[12] would lead to specific and unambiguous legal expressions. Allen's technique is suggestive of two key thoughts: all statements are (1) composed of constituent elements; and (2) built on logical relationships.

He uses implication/co-implication ambiguity[13] to illustrate how symbolic logic could clarify legal imprecision. He considers the conditions for when a seller may rescind a contract or sale as an informative example. Breaking down section 65 of the Uniform Sales Act into six constituent components,[14] Allen argues that even a "relatively simple and straightforward statutory passage…often [has] a wide variety of possible interpretations."[15] For the specific case of section 65, he found that there are eight interpretations a court could take.[16] Yet, of the eight, only one interpretation tends to be adopted by courts, owed to the contextual support of other sections of the statute.

---

[9] *Id.* at 4.

[10] *Id.* at 7.

[11] *Id.* at 8.

[12] Layman E. Allen, *Symbolic Logic: A Razor-Edged Tool for Drafting and Interpreting Legal Documents*, 66 Yale L. J. 833, 835 (1957).

[13] Defined as whether the connection between two elements of a statement is conditional or biconditional. *See* id. at 855.

[14] *Id.*

[15] *Id.* at 857.

Allen suggests, by systematically pulverizing statements of the statute, clearer intentions may be revealed. This method acts as a tool to counter drafting in a "broad and ambiguous form."[17] More recently, Stephen Wolfram made a similar argument. Simplification, he states, could occur through the formulation of a symbolic discourse language. That is, if the "poetry" of natural language could be "crushed" out, one could arrive at legal language that is entirely precise.[18]

Machine-readability[19] appears then to bridge the desire for precision with the inherent logic and ruleness[20] of certain aspects of the law. Machine-consumable legislation may, therefore, be regarded as a product that evolved out of the relationship between syntax, structure, and interpretation. In other words, a potential recipe to resolve the complexity of legalese. What Allen intentionally evades, and is rather significant, is the difference between semantic and syntactic uncertainty. While syntactic uncertainties are often inadvertent, semantic uncertainties are often deliberate. The distinction between syntactic with semantic uncertainty is a mirror to unintentional and intentional ambiguity. This act of categorization implies the capacity to delineate within natural language core tenets of ambiguity.

Therefore, the correlative association between unintentional ambiguity and syntactic uncertainty is an audacious claim that innately reduces the challenges of legislative drafting to a symbolic fix. For now, it appears there may be a stronger argument that symbolic logic is better suited as a metric to assess clarity and precision in legal drafting.

---

[16] Allen conducts a simple mathematical calculation around the number of interpretations. He notes that where the number of antecedents (otherwise, conditional statements) in the statement is equivalent to N, the number of possible interpretations is equivalent to $2^N$. See *id*.

[17] *Id*.

[18] Stephen Wolfram, "Computational Law, Symbolic Discourse, and the AI Constitution," Ed Walters (ed.), *Data-Driven Law: Data Analytics and New Legal Services* 109 (2019).

[19] While there are distinctions in literature between machine-readable and machine-consumable, the author here uses them interchangeably and treats them as synonymous.

[20] Alluding to the quality described in Frederick Schauer, "Ruleness," Dupret Baudouin et al. (eds.) in *Legal Rules in Practice* (2021 Forthcoming).

## B. Plain English Legalese

Marshall McLuhan famously theorized the medium is the message.[21] Developments in law validate this idea. Legalese, much like law itself, is expensive, fragmented, and inaccessible to many who would use it. As such, symptoms of simplification – efforts to make text more digestible – frequently emerge and re-emerge, working through cycles of fashion in the legal industry.

In the 1960s, David Mellinkoff described the absurdity of the legal language bearing characteristics distinct from common speech. Mellinkoff argues that while there is overlap between the two, the language of the law frequently includes common words with uncommon meanings, use of words and expressions with flexible meanings, and "attempts at extreme precision of expression."[22] Perhaps the most interesting is Mellinkoff's sly remarks at the legal language's valiant yet unsuccessful efforts with precision. He notes the contrast between the plays on meaning against the sharp boundaries around the vocabulary. In defense of precision, the arguments often invoked by lawyers is of clarity; that the wording is justified in making the meaning clearer.[23] The cult around precision in law's language has built a fortress around change, projecting a fear that use of plain language would disrupt the clarity associated with legal language.

Therefore, Mellinkoff seeks to debunk this myth of precision; the elusive "exact meaning," desired by lawyers, that keeps the technical language afloat. Alternatively, he finds that the tools used in the legal community do not reflect precision. First, agreement on what is necessarily precise has never been reached.[24] Precision is occasionally defined as being exact or "exactly-the-same-way." The former alludes to a definite term, whereas the latter points at the mechanism of analogy and application of precedent. In either scenario, Mellinkoff finds issue with the understanding of precision. A focus on definite meaning is misleading as legal language often includes vocabulary such as "reasonable," or "substantial" that are fundamentally imprecise.

---

[21] Marshall Mcluhan, *Understanding Media: Extensions of Man* (1964).

[22] David Mellinkoff, *The Language of the Law* 11 (1963).

[23] *Id.* at 292.

[24] Mellinkoff describes this as "the choice of 'precise' language goes by default – without notice that any problem exists." See *id.* at 297.

From the perspective of precedent and argument for tradition, Mellinkoff suggests that precision is merely an effect produced by law's formulas. That is, "an inflexible primitive insistence on word-for-word repetition could make the traditional the precise."[25] Embedded into the legal language is an attachment to form as opposed to meaning. Consequently, the arguments towards precision are, in fact, structural and not linguistic.

Peter Tiersma, decades later, discussed the extent to which legal language was effective as a means of communication. His conclusion was that the goals of the language did not serve the intentions of the law. That is, the desire to appear objective and authoritative conflicted with the use of language in law. Tiersma suggests that legal language has come to be understood as a method of exclusion, an indicator that one belongs to a "legal fraternity."[26] This incongruency enables a continued dependence on the legal community to decipher and translate legal texts.

Tiersma highlights two elements that have worked against the use of plain English in law: (1) the "quest for precision" in law; and (2) the legal lexicon. The former acts as a shield against ordinary English, and the latter is to distinguish law from other disciplines. Perhaps ironically, Tiersma observes that the arguments for legal language – clarity, conciseness, and precision – are also the causes of imprecision and lack of clarity. Like Mellinkoff, he argues that the legal language strategically plays on imprecision, flexibility, and generality of use, as well as a specific vocabulary that is largely arcane and jargon.[27] Moreover, interpretation plays a different function in legal than in ordinary language. Tiersma suggests that in ordinary English, interpretation is focused on the speaker's meaning. In legal interpretation, it is fundamentally a semantic exercise reinforced by the aforementioned lexicon. The differences in the practice of language and the reasons behind their use, in effect, lead to complications surrounding the inclusion of plain English in law. Consequently, decades of effort in converting complex legal language to plain English have been met with minimal success.[28]

---

[25] *Id.* at 299.

[26] Peter Tiersma, *Legal Language* (1999), available at: http://languageandlaw.org/LEGALLANG/LEGALLANG.HTM

[27] *Id.*

[28] In addition to the ongoing dialogue towards a 'plain legal English,' it is perhaps best summarized by William Pitt on the elusiveness and illusion of achieving this conversion. *See* William Pitt, "Fighting Legalese with Digital,

Nevertheless, there have been strong efforts of developing a plain English for the legal community. Richard C. Wydick, inspired by Mellinkoff, addresses the design problem raised by Tiersma. The underlying argument is that "good legal writing is plain English."[29] Wydick suggests that distinguishing a legal from ordinary language hinders, rather than promotes, legal work. Furthermore, he contends that there are several quick fixes to translating existing legal to plain language. In his text, Wydick identifies issues of legal language as semantic ones of choice and arrangement. The central discussion is on word use and how to manipulate them "with care."[30] Grammar is equally relevant; to consider foremost the active voice and punctuation.

There have been examples of Wydick's suggestions in practice. The Plain English Movement[31] reflected an eager intent to increase the accessibility of legal knowledge to those outside of the legal community. This was owed to the rising demand for important consumer documents to be made understandable to the general population.[32] Similarly, this has permeated into calls for plain English legislation. Guidelines of 'good faith' were written for legislation to use active verbs and short sentences and be capable of passing the Flesch test.[33]

Despite the vast improvements to the language of consumer documents, most legal documents continued to be written in legalese. If the shift from legal to plain English is as simple and intuitive as described by Wydick, the question becomes: why have the peculiarities of legal language and drafting, persisted? In line with Tiersma's suggestion, perhaps it may be a result of exclusivity. That is, the complexity of the language fosters a continual reliance on the legal community, reinforcing the need for a knowledge translator. On the other hand, there may be a more subtle reason for the preservation of legalese. This argument draws from Mellinkoff's discussion of tradition. Provided that legal language has always been housed in a particular form, there rests an underlying hesitation that legal concepts cannot be expressed in another way. Though

---

Personalized Contracts," Harvard Business Review (February 27, 2019), https://hbr.org/2019/02/fighting-legalese-with-digital-personalized-contracts.

[29] Richard C. Wydick, *Plain English for Lawyers* (2005).

[30] *Id.*, see importantly chapters 6 and 7.

[31] This began with revisions around promissory notes introduced by Citibank in the 1970s. *See* Tiersma, *supra* 26.

[32] *Id.*

[33] This was considered a "readability" assessment, as it measures the average length of sentences and words. It was suggested that this acted as an objective and quantifiable measurement for comprehensibility. See *id.*

Mellinkoff ascribes this to the illusion of precision, it may in fact be an inability to reconceptualize the law. This would imply a marriage to the form. In this case, enabling machine-readability would demand perpetuating existing forms of legal expression.

### C. Why don't we layer it? XML in Law

From plain English, there took a technical turn. Building on the previous innovations to the technical infrastructure of law, there are new opportunities enabled by the digital and the computational. As demonstrated by the framework for licenses introduced by Creative Commons, contracts can be structured in order to standardize legalese, develop human readable summaries of complex ideas, and integrate them into a machine-readable format.[34]

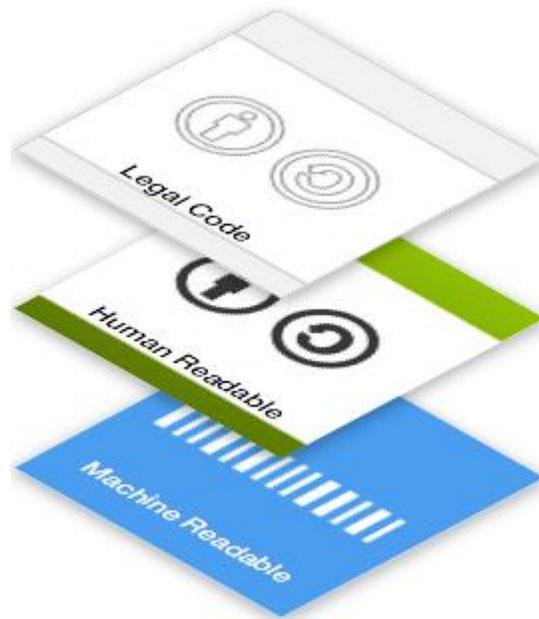

35

In hopes of developing a better understanding of legislative documents, LegalXML and LegalDocumentXML, products of OASIS Open,[36] were created to provide a common legal document standard "for their interchange between institutions anywhere in the world and for the creation of a common data and metadata model that allows experience, expertise, and tools to be shared and extended."[37]  This standard-based approach  focuses on assessing the ways in which machine-readable information may be integrated into the official text of legislative documents.[38]

---

[34] "About the Licenses," Creative Commons (accessed Jul. 28, 2021), https://creativecommons.org/licenses/.

[35] "Layers," Creative Commons (accessed Jul. 28, 2021), https://creativecommons.org/share-your-work/licensing-considerations/layers/.

[36] OASIS Open (accessed Jun. 12, 2021), https://www.oasis-open.org/.

For a document to be made machine-readable, a descriptive markup meta-language,[39] like eXtensible Markup Language (XML), must be embedded into the text in order for a computer to understand it. That is, the document must be deconstructed and sorted into components based on structure and semantics. Structure is defined as the organization and categorization of various parts of the document on the basis of functionality.[40] Semantics, on the other hand, is defined as the meaning, or what the information within the document represents. The intention, then, of decomposing documents into respective structural and semantic framings enables developing a taxonomy and ontology around organizing legislative information.

In effect, standardization is an argument for drawing out and weaving similarities between legislative documents across various jurisdictions. The aim is to increase accessibility and fortify interoperability within the legal ecosystem.[41] As opposed to the existing ad-hoc, or piecemeal, method, the application of a standard technique would encourage transparency in the production and dissemination of legislative information.

As an initial response to a United Nations project to strengthen information systems in legislatures in Africa, a set of standards and guidelines for digital Parliament services, known as the Architecture for Knowledge-Oriented Management of Any Normative Texts using Open Standards and Ontologies (Akoma-Ntoso), was developed.[42] This framework sought to manage information and recommend technical policies and specifications for building Parliament information systems.[43] The results of Akoma-Ntoso led to the three key achievements: (1) the Akoma-Ntoso XML schema; (2) a labelling convention for legal resources (URI); and (3) Legislative Drafting Guidelines.[44] These achievements reflect the broader vision on the use of XML to provide a stronger structural and semantic framework around organizing parliamentary and legislative information. The Akoma-Ntoso XML schema (Akoma-Ntoso), in particular, enables the inclusion of descriptive structure to the content of legislative documents; and, thereby, providing context to legislative information.[45]

---

[37] OASIS LegalDocumentML (accessed Jun. 12, 2021), https://www.oasis-open.org/committees/tc_home.php?wg_abbrev=legaldocml

[38] Fabio Vitali, "A Standard-Based Approach for the Management of Legislative Documents," Giovanni Sartor et. al (eds), *Legislative XML for the Semantic Web* (2011).

[39] A form of language used in web programming to allow users to identify individual elements of a document. See lecture slides, "Web Programming," https://home.adelphi.edu/~siegfried/cs390/390l6.pdf.

[40] Vitali, *supra* 38 at 39.

[41] *Id.* at 38-42.

[42] Monica Palmirani and Fabio Vitali, "Akoma-Ntoso for Legal Documents," Giovanni Sartor et. al (eds.), *Legislative XML for the Semantic Web* (2011).

The Akoma-Ntoso architecture has been revered as the bedrock on which LegalXML is built.[46] There are two key principles that are fundamental to the schema: (1) descriptiveness; and (2) prescriptiveness. The former emphasizes the preservation of the original "descriptiveness" of the document. This suggests that there is no loss in the integrity of the legislative document, specifically qualitative components that provide important legal or regulatory context. The latter focuses on the implementation of rules, "directly drawn from the legal domain."[47] Together, these principles imply and, perhaps, reaffirm the notion that it may be possible to sort within legal documents elements that are inherently executable and structured; and others that require the detail and particularity. More importantly, Akoma-Ntoso places a focus on the representation and validity of legal documents.[48] The design purports to place at the forefront a proper reflection of legal concepts.

Monica Palmirani and Fabio Vitali describe four generations of LegalXML, with Akoma-Ntoso understood as the third generation.[49] Though the differences between generations is primarily based on nuances of structuring, the third generation onward relies on a thorough understanding of object-oriented design.[50] That is, an assessment of patterns and classifications are coupled with an analysis of the relationships between text, structure, and metadata. This process is central to the schema and translation of legal concepts.

---

[43] *Id.* at 75.

[44] *Id.*

[45] *Id.* at 76.

[46] *Id.*

[47] *Id.* at 77.

[48] *Id.*

[49] *Id.* at 78.

In effect, the third generation establishes the "complex multilayered information architecture"[51] that decomposes the legal document from pure text to structured analysis. This multilevel construction is described as a semantic web layer cake.[52] Modelling the document into layers, text and structure, metadata and ontology, aligns again with the implied argument that the content of legislative documents are innately categorical. That is, as opposed to a reconfiguration, or a reframing, of the document, it is instead a question of rearrangement and extraction of these structured elements.

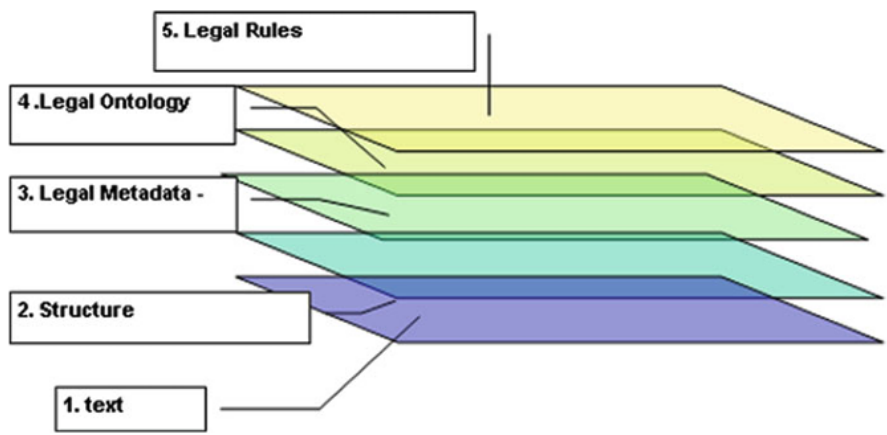

**Fig. 6.1** Layers of representation in Legal Document Modelling

How then does LegalXML work? Below are examples[53] of how the layers are drafted in Akoma-Ntoso XML schema and how the relationships between these layers operate. Beginning with the text and structure layers, both layers take from the original natural language and annotate each element semantically. As notable in the examples, the text and structural markup (denoted by these parameters </>), indicate to the machine how the document is organized. Textually, it highlights between paragraphs and references. Structurally, it highlights headers, sections, and subsections.

---

[50] *See* for example in Megan Ma et al., *Deconstructing Legal Text: Object-Oriented Design in Legal Adjudication*, MIT Computational Law Report Release 1.3 (2020) available at: https://law.mit.edu/pub/deconstructinglegaltext/release/1.

[51] Palmirani and Vitali, *supra* 42 at 78.

[52] *Id.* 79.

[53] All examples are taken directly from Palmirani and Vitali's demonstration in their article.

**Fig. 6.2** Example of text markup

**Fig. 6.3** Example of structure markup in Akoma-Ntoso

At the metadata layer, annotations become more complex. As opposed to indicating a legislative document's logical connectors and organization, metadata represents the interpretation and context of the document. In the example below, the left panel of the screen represents a textual markup of a particular section of legislation. The right panel reveals the underlying possibility for multiple interpretations of this section. Therefore, the <mod id=mod1> denotes that for this specific case, there may be two equally valid interpretations: (1) authentic; or (2) exception.[54]

---

[54] *Id.* at 82.

**Fig. 6.4** Example of metadata markup connected to the structured text

Moreover, metadata annotations clarify the "local" meaning.[55] For reasons of simplification and uniformity across categorization, Akoma-Ntoso intentionally uses a single convention for all documents. This enables a "shared conceptual architecture"[56] across the legal ecosystem. Therefore, to avoid confusion, the metadata annotates the specific meaning at hand. Below, the docProponent refers to the source of authority. In the left panel, the legislation indicates the legal authority draws from the Ministry of Local Government. The right panel indicates the source draws from the Supreme Court of Appeal.

---

[55] *Id.*

[56] *Id.*

```
<preface>                           <header>

  <p class="heading">REPUBLIC OF       <p>
AMCE</p>
                                       <b><docProponent refersTo="#SCOA">
  <p class="subheading">             SUPREME COURT OF APPEAL
                                     ACME</docProponent></b>
   <docTitle> GOVERNANCE FRAMEWORK
BILL</docTitle>                        </p>

  </p>                              </header>

  <p class="subheading">

   <docProponent>MINISTER FOR LOCAL
GOVERNMENT of ACME

   </docProponent>)

  </p>

</preface>
```

**Fig. 6.5** Example of shared elements with different semantic meanings

Equally, this shared vocabulary behaves as a legal ontology. It indicates how components of legislative documents belong to broader categories within a legal ecosystem. In the aforementioned, the metadata annotations reveal how a particular piece of legislation connects with other legal documents. More importantly, it localizes where specific interpretations are drawn. This substantiates a more explicit approach on the gathering and understanding of legal knowledge.

Akoma-Ntoso then fulfills the desires of logicians for a legal language that is sufficiently precise. Returning to Allen, if legislation should have a clear structure, Akoma-Ntoso appears as an ideal option. Yet, the rate of its adoption has been strikingly low.[57] This is perhaps owed to the two-fold complexity of migrating legislative documents from text to XML and the requirement of XML competency in the translation process. First, converting legislation from natural language to an XML schema is described as an eight-step recipe.[58] Importantly, it requires first a legal analysis that is typically done on paper. As described by Palmirani and Vitali, the legal expert must meticulously and manually conduct the process – sorting within legal documents the text, structure, metadata, and ontology. As well, the legal expert must be fluent in Akoma-Ntoso, correctly annotating the elements and identifying the legal relationships latent in the documents.

---

[57] "Use Cases," Akoma Ntoso, available at: http://www.akomantoso.org/?page_id=275.

[58] Palmirani and Vitali describe in further detail the process of taking text and structuring. *See* Palmirani and Vitali, *supra* 42 at 94-98.

In effect, though Akoma-Ntoso offers benefits of making legal language machine-readable and preserves the richness of legal concepts, its use requires significant costs. The process is rather laborious, and few legal experts[59] currently have the technical skills to draft in XML schema. Consequently, this has contributed to rather lackluster enthusiasm for its adoption.

### D.  Old Wine in New Bottles: Rules as Code

Still, machine-readable legislation has received renewed popularity. This is perhaps owed to the release of the recent OECD Observatory of Public Sector Innovation Report titled, "Cracking the Code: Rulemaking for Humans and Machines" (OECD Report).  The OECD Report articulates how machine-consumable, defined as machines understanding and actioning rules consistently, reduces the need for individual interpretation and translation[60] and "helps ensure the implementation better matches the original intent."[61] This methodology enables the government to produce logic expressed as a conceptual model – in effect, a blueprint of the legislation.

These ideas are reminiscent of Anthony Casey and Anthony Niblett's thought experiment on the micro-directive.[62] Interestingly, one of the underlying fascinations with Rules as Code lies in the types of statutes subject to digital transformation. Rules as Code applies two general practices of 'code-ification': (1) programming tasks; and (2) knowledge-based systems. The former is more direct, while the latter poses epistemic challenges. Programming tasks may be defined as a legislative calculator; the legal questions asked are already known and understood in advance. Typically, these tools are designed to assess eligibility, particularly in the fields of taxation and benefits law. OpenFisca, the most widely known example, is an open-source platform that writes rules as code. The available code focuses on legislation that "can be expressed as an arithmetic operation."[63]

---

[59] It must be noted that the XML vocabulary and schemas are open-source and publicly available. This suggests that while the documentation is available, it continues to remain limited amongst those willing to adopt the practice. See for example, OASIS LegalDocumentML, *supra* 37.

[60] OECD Observatory of Public Sector Innovation, *Cracking the Code: Rulemaking for Humans and Machines* 19 (2020).

[61] *Id.* at 22.

[62] To recall, in this futuristic construct, lawmakers would only be required to set general policy objectives. Machines would bear the responsibility to examine its application in all possible contexts, creating a depository of legal rules that best achieve such an objective. The legal rules generated would then be converted into micro-directives that subsequently regulate how actors should comply with the law. *See* Anthony J. Casey & Anthony Niblett, *The Death of Rules and Standards*, Coase-Sandor Working Paper Series in Law and Economics No. 738 (2015).

Knowledge-based systems, on the other hand, encode rules required to arrive at a specific legal question. That is, these tools consist of logical algorithms that help identify the legal knowledge to be gathered from a particular statute. They come from the lineage of expert systems and logic programming. DataLex Knowledge-Base Development Tools (DataLex), for instance, is a rules-based legal inferencing platform that draws, from legislative texts, conclusions based on antecedents. In effect, the DataLex software is powered on propositional logic.[64]

Despite differences between practices of 'code-ification,' the types of legislation amenable to a Rules as Code approach predicate on an inherently mathematical structure. This suggests that for legislation with clear formulaic rules, expression in symbolic logic is intrinsically available. Ruleness becomes the essential ingredient. The OECD Report, however, does not distinguish between types of legislation and, rather, conflates legislation under a seemingly uniform banner.

Though the OECD Report succeeded in providing a comprehensive overview of Rules as Code, there remains a gap around the practical implementation and the form machine-readable legislation should take. The OECD Report anticipates three approaches to building machine-consumable legislation: (1) a manual coding of the legislation across a multidisciplinary team; (2) the use of semantic technologies; and (3) a domain model-based regulation, whereby the government would create an official model of rules to then convert to software languages.[65] These approaches drew inspiration from a deeper analysis on the levels of digitization.[66] Unlike Meng Weng Wong's aspirational vision for machine-readability, the OECD Report is agnostic to these possible methods.

Recent implementations of Rules as Code have surfaced globally. Currently, the most prominent example is found in Australia. In the summer of 2020, the New South Wales (NSW) Government released its first Rules as Code legislation to reduce ambiguity and simplify interpretation.[67] Built on the OpenFisca platform,[68] the *Community Gaming Regulation 2020* (Gaming Regulation) identifies "the conditions for running community games by different charities, not-for-profits and businesses in NSW."[69] The *Gaming Regulation* is drafted in several forms: machine-readable, human readable, and on a computing interface. Perhaps its most incredible achievement is the publicly available digital version of the *Gaming Regulation*. The NSW Fair Trading website enables those engaging with the regulation to determine whether their activity is permissible and if an authority is required to conduct the activity.[70] This website is considered a "single source of truth" that will increase transparency and efficiency, by reducing time spent understanding the regulation, and providing easily digestible responses to particular situations of concern.[71] The website offers information on various sections of the legislation in plain language. The prize jewel, however, is its questionnaire.

---

[63] "Before You Start", Open Fisca Documentation (accessed January 2021) https://openfisca.org/doc/. For further details on how to 'translate' from law to code, see: https://openfisca.org/doc/coding-the-legislation/index.html.

[64] "Legal Inferencing Systems: Supporting provision of free legal advisory services," DataLex (accessed January 2021) http://austlii.community/foswiki/pub/DataLex/WebHome/DataLex_intro.pdf.

[65] OECD Observatory of Public Sector Innovation, *supra* 60 at 63-66.

[66] Meng Weng Wong, *Rules as Code – Seven Levels of Digitisation*, Research Collection School of Law (2020).

In experimenting with the website's questionnaire, the "Community Gaming Check,"[72] the key content behind the legislation appears to be logically reducible and fundamentally arithmetic. Below are two sample snapshots of completed questionnaires:

---

[67] "In an Australian first, NSW is translating rules as code to make compliance easy," NSW government digital.nsw (accessed Jun. 12, 2021), https://www.digital.nsw.gov.au/success-stories/australian-first-nsw-translating-rules-code-make-compliance-easy.

[68] To see the regulation housed on the OpenFisca platform, *see* Openfisca-Nsw-Base Web API (accessed Jun. 12, 2021) http://nsw-rules-dev.herokuapp.com/swagger.

[69] *Id.*

[70] *Id.*

[71] *Id.*

[72] For further detail and/or to experiment with the questionnaire, *see* "Community gaming check," NSW Government Fair Trading (accessed Jun. 12, 2021), https://www.fairtrading.nsw.gov.au/community-gaming/community-gaming-regulation-check. For the machine-readable version of the legislation, *see* Openfisca-NSW (accessed May 10, 2021) https://github.com/Openfisca-NSW/openfisca_nsw_community_gaming.

**Can I conduct my gaming activity?**

← Go back

**You may run this gaming activity without an Authority**

More information can be found on the **Community gaming** page.

**What you've answered**

| | |
|---|---|
| 1. Type of game | **Free lottery** |
| 2. Total prize value of all prizes from gaming activity | **$1000** |
| 3. Free participation | **Yes** |
| 4. Prize consists of money | **No** |

**You may not run this gaming activity**

More information can be found on the **Community gaming** page.

**What you've answered**

| | |
|---|---|
| 1. Type of game | **Promotional raffle** |
| 2. Gaming activity on authority of reg club | **Yes** |
| 3. Venue is registered club | **Yes** |
| 4. Gaming activity organised for patronage | **Yes** |
| 5. Gross proceeds from gaming activity | **$3000** |
| 6. Proceeds used for meeting cost of prizes | **$200** |
| 7. Total prize value from single gaming session | **$200** |
| 8. Prize consists of money | **No** |

Presumably, for the purposes of simplification, the questions are either drafted in binary or are numerically driven. As a result, the Community Game Check (CGC) will compute a response in the affirmative or negative. The underlying assumption of the CGC is that the legislation raises one of two questions: (1) determining whether a community game is admissible; or (2) if authority is required. Again, it may be reaffirmed that Rules as Code focuses on prescription and rules; description continues to fall within the jurisdiction of the original natural language version. Underlying this focus is the assumption that legislation is largely mathematical and that legislative questions may be solved through predicate logic.

Alternatively, the Rules as Code initiative sparked more granular innovations, including formal languages compatible for its drafting and expression. Catala, "a new programming language created by lawyers and computer scientists for quantitative statute formalization,"[73] is a proposed solution for computing tax and benefits legislation. In their article, Denis Merigoux and Liane Huttner explore the issues of existing expert systems used for tax and benefits law. They first outline that the use of antiquated code – programming languages that "exceeded the tenure of its original programmers"[74] – risks the inability of adapting to new functional demands. This has evident ramifications provided the evolving nature of legislation. Equally, they explore the pitfalls of using existing algorithmic tools for tax collection that has led to both miscalculations and barriers with revision.[75]

---

[73] Denis Merigoux and Liane Huttner, *Catala: Moving Towards the Future of Legal Expert Systems*, HAL archives-ouvertes (2020).

[74] *Id*. at 2.

Their recommendation is to use formal methods coupled with literate pair programming in order to tackle the aforementioned issues. First, literate pair programming is a hybridized understanding of literate and pair programming in software development.[76] Merigoux and Huttner suggest that a combination of these methods, and between a lawyer and computer scientist, enable quality assurance in the translation of law to code. The line-by-line annotation of statutory texts allows for a "local discussion" on the "lawful interpretations of the statutes."[77] Evidently, this recommendation aligns closely with one of the OECD Report's anticipated approaches to building machine-consumable legislation: a manual coding of the legislation across a multidisciplinary team. However, the more pressing question is the use of formal methods.

Formal methods are a restructuring of abstract concepts to "mathematical objects."[78] Formal methods act as mathematical proofs, determining functional equivalence.[79] Effectively, it is reminiscent of Carnap's logical syntax and treatment of language as a calculus. As a result, this practice depends on the existing and inherent formal structure of the legislation.[80] This again reinforces the requirement of ruleness in Rules as Code. Consequently, while Merigoux and Huttner's recommendations ensure that legal quality is maintained, Catala's benefits remain within the limited scope of intrinsically quantifiable legislation.

Looking outside law again for insight into how it might be possible to describe meaning consistently to the language of law, there are more opportunities for insight. This approach of abstracting formal texts into different layers that might be understood in different circumstances is in line with Karl Friston's work on contemporary work in computational neuroscience, i.e. the formal methods for understanding and distinguishing between the existence and non-existence of observed phenomena as represented in a Makrov blanket.[81] So while the medium of law informs its message, ultimately the broader understanding of the conceptual boundaries of law, even from a neuroscientific level, needs to be represented by accounting for different of the various active and sensory states that are observed when seeking to answer questions like "what is a rule?"

---

[75] *Id.* at 3.

[76] Literate programming is described as line-by-line annotations, while pair programming is pairing two programmers in the production of code. For further detail, see *id.* at 7.

[77] *Id.*

[78] *Id.* at 6.

[79] *Id.*

[80] Merigoux and Huttner state explicitly the assumption of expression in mathematical terms as well as the "formal specification" of statutes. See *id.*

### E. Legislative Tinkering

Recent implementations of Rules as Code fortify the argument that, currently, machine-consumable legislation is limited to highly structured legislation. Nevertheless, these examples leave one question fundamentally unanswered: *what should be the role of machine-readable legislation?* Is it simply a 'coded' version of the legislation; or is it a parallel alternative, one that is legally authoritative? Or is it a domain model of regulation from which third parties derive their own versions, akin to an open-source code? These three scenarios have their own sets of implications. Only in clarifying the role of machine-readable legislation would a fruitful assessment of how logic syntax and symbolic language are capable of representing legal knowledge.

#### i. Authoritative Conundrum

New Zealand released in March 2021 its own version of the OECD Report, "Legislation as Code for New Zealand: Opportunities, Risks, and Recommendations" (Legislation as Code Report). One of the key conclusions of the report calls for a distinction between competence and desire. That is, even if legislation may be drafted in code, it should not be. Unlike the OECD Report, the Legislation as Code Report takes a strong stance on the role of machine-consumable legislation. The report argues that rules drafted in code "should remain subordinate to legislation," stating that "enacting code creates serious constitutional confusion and risks undermining the separation of powers."[82]

---

[81] Karl Friston, *A Free Energy Principle for a Particular Physics* (2019). Available at: https://arxiv.org/pdf/1906.10184.pdf

[82] New Zealand Law Foundation Law and Information Policy Project, *Legislation as Code for New Zealand: Opportunities, Risks, and Recommendations* 3 (2021).

This is owed to the law's "technological use of written natural language;" whereby the use and interpretation of words keeps in balance the structure of the law with its institutions. [83] As code does not have the same interpretive space as natural language, this runs the risk of the judiciary being unable to perform its constitutional role relative to statutory interpretation.[84] Accordingly, the inability to invalidate legislation for inconsistency, given interpretative barriers with code, would "degrade the rule of law."[85]

ii.     *Language Shopping*

The Legislation as Code Report further contrasts the OECD Report by concluding that parallel drafting is not a solution, but simply a mitigator to issues of interpretation.[86] Provided that perfect translation does not exist, there is inevitably potential for meaning to diverge even if a common intent is established. Therefore, while an encoded version arguably reflects *an* interpretation of the law,[87] machine-consumable legislation that has legal authority raises, equally, issues analogous to both legislative bilingualism and bijuralism.[88] This could foreseeably create statutes with multiple personalities, having dissonance between linguistic variants and heightening ambiguity in interpretation.

In this regard, Canada is an informative example. In 1995, the formal adoption of legislative bijuralism led to an acknowledgment of four legal audiences in Canada; that there is a "right to read federal legislation in the official language of their choice and to find that legislation terminology and wording [to be] consistent with the system of private law in effect in their province or territory."[89] As such, the constitutional requirement for all legislation to be written bilingually forcibly produced makeshift equivalents in legislation, devised without standard nor appropriate concern for the problems of interpretation.

---

[83] *Id*. at 9.

[84] *Id*. at 58.

[85] *Id*.

[86] *Id*. at 4.

[87] In fact, the Legislation as Code Report suggests that it may be useful to focus on the opportunities for approaches of non-authoritive implementations of Rules as Code. See *id*. at 9.

[88] Lionel A. Levert, "Harmonization and Dissonance: Language and Law in Canada and Europe," Department of Justice Canada, *Bijuralism and Harmonization: Genesis* (May 7, 1999) https://www.justice.gc.ca/eng/rp-pr/csj-sjc/harmonization/hfl-hlf/b1-f1/bf1e.html.

[89] *Id*.

There are two models of producing bilingual legislation: translation and co-drafting. While they are perceived as distinct, the process around crafting bilingual legislation often involves a hybridization of both. This typically results in a conceptual mismatch between one language to the other. Michael J. B. Wood provides a fascinating illustration through the word 'any.'[90]

| | |
|---|---|
| **(1)** The report shall include **any** document specified in the schedule. | **(1)** La rapport comprend **l'un des** documents énoncés à l'annexe.<br>**(1)** Le rapport comprend **les** documents énoncés à l'annexe. |

In the English language, 'any,' in the affirmative, describes 'one' out of a specific list. In the above example, the intention of the drafter may be to indicate that, should there be documents specified in the schedule, they should be included. However, to the reader, it may suggest that any one of the documents specified in the schedule should be included. Consequently, in the French language example, there produces two variants. This lack of equivalence in the word 'any' produces ambiguity between versions of the legislation. Both of which have equal authority under Canadian law. Wood discusses other examples including pronominal phrases such as 'thereof' and chains of qualifiers.[91] In the former, phrases of this type often foster confusion, particularly in co-referencing.[92] As well, there are no direct equivalents in French. In the latter, the Germanic origins of the English language allow nouns and adjectives to be chained together. This use of grammar does not exist in French. Instead, the French language applies a series of modifying phrases. Consequently, if meaning is unclear and ambiguous in English, there is potential for further complication in French.[93]

---

[90] Michael J.B. Wood, *Drafting Bilingual Legislation in Canada: Examples of Beneficial Cross-Pollination between Two Language Versions*, 17 Statute. L. Rev 66, 70 (1996).

[91] *Id*. at 70-72.

[92] *See* example on "a part thereof." *Id*. at 70.

[93] *Id*. at 71-72.

Likewise, the presence of both civil and common law systems within Canada has led to complications with the translatability of legal concepts. Bijuralism stipulates the requirement to have proper terminology and notions present across both systems of private law in Canada. To achieve this requirement, the most frequent methods used are the "neutrality technique" and the "doublet."[94] The former is simply the use of 'neutral' terms or phrases in defining concepts without particular connection to either one of the systems. The latter is to enable the co-existence of legal concepts when there is no functional equivalence. In cases of the doublet, both versions of the legislation "retain their separate identities."[95] This means that paragraphs within the same legislation may have intentional signposts to direct how the rule of law is to be applied depending on the system.[96] Typically, both expressions of the legal concept appear one after the other in each language version.

Evidently, problems of interpretation arise as "civil law terms are juxtaposed with common law expressions."[97] Within the country, there were issues symptomatic of conflict of laws; whereby courts applied common law definitions to jurisdictions that followed civil law systems. This led to inconsistencies in precedent, as civil and common law terminology were used interchangeably without proper regard for the nuances of legality between each system's interpretations.

Canada has since made remarkable strides in legislative bilingualism and bijuralism. This was owed to a reframing of federal requirements as a strain of comparative law, as well as the subsequent emergence of jurilinguists; otherwise, experts trained in both systems.[98] Returning to machine-readability and authoritative code, what are some lessons that can be drawn from the Canadian experience? First, there has been a rise in interdisciplinary training between law and computer science. Mireille Hildebrandt's recent textbook is a prime example. *Law for Computer Scientists and Other Folk*, as she describes, is an endeavor to "bridge the disciplinary gaps" and "present a reasonably coherent picture of the vocabulary and grammar of modern positivist law."[99] As well, law schools are beginning to offer technology and innovation courses including training in computer programming.[100] This is facilitating a growth and demand in experts fluent in both disciplines.[101] Moreover, as evidenced, co-drafting can be seen in the recommendations and development of machine-readable languages like Catala.

---

[94] Levert, *supra* 88.

[95] *Id.*

[96] *Id.*

[97] *Id.*

[98] Universities of Ottawa and departments of jurilinguistics produced both common law terminology in French and civil law terminology in English. This pioneering work offered the potential to better capture the necessary distinctions and comparisons between the two systems of law. See *id.*

There remains, however, a significant gap in both reconciling and harmonizing legal concepts between code and natural language. Perhaps the deeper question is whether and how that may be possible. In Canada, common and civil law terminology come from existing traditions of private law. Their respective expressions are rooted in legal history. However, there is neither a comparable legal system nor a comparative field of law for code. That is, code could only potentially extend as an alternative language, but not as a system of norms. The functional limitations of code could only be interpreted as linguistic limits, whereas normative principles of programming and computer science could never be perceived as parallel legal principles. As a result, the discussion raised in the Legislation as Code Report, on the risk of authoritative code degrading the rule of law, is a critique of code as a legal mechanism. The complexity lies in the extent to which the linguistic medium has the capacity to alter the integrity and character of the law, even if the intention of its use is simply expression.

###### iii.    *The Alchemy of Legal Architecture*

Perhaps the most understated challenge with Rules as Code hinges on the legal infrastructure.[102] Across several possible approaches to machine-readable legislation, there remains unresolved questions of design and interoperability between legal documents. That is, if a new symbolic language, like code, effectively enforces a controlled grammar, what are its implications as it moves across the legal ecosystem; in particular, its interactions with various legal sources?

---

[99] Mireille Hildebrandt, *Law for Computer Scientists and Other Folk* (forthcoming OUP, 2020). A web version is currently accessible on the open-source platform: https://lawforcomputerscientists.pubpub.org/ .

[100] Law schools are beginning to offer courses in technical development, including computer programming. Moreover, classes that apply design-thinking to legal studies and were developed with the intention of acknowledging technology as a powerful driving force in law. Consider Harvard Law School and Georgetown Law School's Computer Programming for Lawyers classes, or Innovation Labs at Northwestern Law School or The Design Lab at Stanford Law School. See for example Harvard Law School, *Computer Programming for Lawyers* (accessed February 2020), https://hls.harvard.edu/academics/curriculum/catalog/default.aspx?o=75487.

[101] "Embedded technical expertise may be necessary to design, develop, and maintain useful and useable tools," also "development of the tool resulted from a multi-year strategic plan to hire lawyers with coding skills…" See David Freeman Engstrom and Daniel E. Ho, "Artificially Intelligent Government: A Review and Agenda" in Roland Vogl (ed.), *Big Data Law* (2020).

[102] Gillian Hadfield, *Rules for a Flat World* (2016).

Reflecting back on the Legislation as Code Report, one important argument raised is the acknowledgment of legislation as "one component among many that comprise the wider system of laws and rules."[103] Statutes frequently reference one another, highlighting a "process of synthesizing multiple inputs into a contextually dependent output."[104] Provided that legislation are not perceivably independent texts, it is then important to consider how machine-readable legislation works in tandem with other legal documents.

In the OECD Report, the discussed approach for a domain model-based regulation is one that raises persistent queries on interoperability. Should there be a government-endorsed model from which legislation will be converted into third-party machine-readable versions, this could create inconsistent interpretations; thereby, testing the legal limits of the model. Currently, there is no standard for how the model translates to individual policies. More importantly, what might be issues of fit between various machine-readable documents, such as between machine-readable legislation to machine-readable contracts?

In late December 2020, the University of Cambridge announced the launch of the Regulatory Genome Project.[105] As opposed to legislation, the focus of the project is on regulation, and specifically financial regulation. The Regulatory Genome Project intentionally steers away from regulation as code and considers the notion of "sequencing."[106] Rather than translation, regulatory information will be extracted and placed in a data repository. The regulatory data will then be organized into a taxonomy. In accordance with the taxonomy, experts will annotate key information and build a training set. This model will then be used to subsequently generate machine-readable regulatory documents. In effect, it is a process of retrieving the contents of regulation from an openly accessible platform that bears a specific framework of capturing the regulatory data. This permits a single source of 'truth' and a common standard for accessing machine-readable regulatory information.

---

[103] Legislation as Code Report, *supra* 82 at 48.

[104] *Id.*, at 50.

[105] "The University of Cambridge announces the launch of the Regulatory Genome Project to sequence the world's regulatory text through machine learning," *The Regulatory Genome,* https://regulatorygenome.com/news/university-cambridge-regulatory-genome-project/.

[106] The Regulatory Genome Project, *The Regulatory Genome* (accessed March 10, 2021) https://regulatorygenome.com/about-us/.

The significance of this approach is its departure from language design. That is, as opposed to dwelling on the semantic conversion of natural language to code, the project turns its attention to the information contained in regulation. It is simply a complete rewrite, or paradigm shift, of digesting regulation. Beyond an interdisciplinary collaboration, the Regulatory Genome Project has received the support of regulators, authoritative figures of the community, "to validate and refine the taxonomies to enable effective benchmarking across jurisdictions globally."[107] Interestingly, this parallels an amalgam of the Rules as Code domain-model with the Legislation as Code argument that the variability of interpretations would be limited if authoritative interpretations are made available.[108]

As a result, the Regulatory Genome Project offers an unconventional method for machine-readability. Evidently, this may be simpler with regulation than it is with legislation. Namely, legal authority operates differently than regulatory authority. In considering this approach, the challenge would be systemic and one that requires convincing a complex network of legislative and judicial power to construct laws on an entirely separate paradigm. Nonetheless, it offers a perspective on mediums of communication and computational modelling that extends beyond language to a level of further granularity: data.

Existing literature has focused on the promise of Rules as Code as the magical formula for increased clarity and precision in legislative drafting. Undeniably, machine-readable legislation has deep-seated roots in logical syntax and symbolic language. The Legislation as Code Report, however, highlights that further discussion is required in better defining both the legal function and status of machine-consumable legislation. Fundamentally, machine-readable legislation

---

[107] *Id.*

[108] Legislation as Code Report, *supra* 82 at 82.

requires a space for judicial and legal contest; effectively, an appeal process in the event of dispute.[109]

This is not to say there is no place for machine-readable legislation. In fact, the Legislation as Code Report argues that computational models can be commendable if the model is (1) "legally correct," and (2) there is infrastructure in place "to assess how the law has been interpreted and modelled."[110] For example, the Legislation as Code Report cites the Auckland District Law Society's Standard Form Agreement for Sale and Purchase of Real Estate (ADLS Standard Form). The ADLS Standard Form is described as an instrument that "embod[ies] a reliable interpretation of multiple primary legal sources" and "indicate[s] the value that similar interpretation might have if they are coded and modelled reliably, while retaining the ability to scrutinize them through legal argument."[111] Provided that this agreement has been drafted and revised within a dependable legal environment, the ADLS Standard Form has demonstrated the potential for reproducibility while maintaining certainty. This suggests that finding existing natural language documents with an accepted standard and structure may be appropriate for computational modelling.[112] Again, this reinforces that Rules as Code is available only in narrow-use cases, specifically, legislation with inherent logical structures.

At a broader epistemological level, there remains limitations from the perspective of knowledge representation; in turn, forcibly demanding a reflection on the intentions and purpose of laws. The Regulatory Genome Project has revealed that there may be an alternate option of consuming information. As law has language at its core, interpretation has centered on the linguistic exercise. This has led to a heavy reliance on translation when reconciling human with machine-readability. However, lessons from core linguistics suggest that natural language is composed of three underlying components: syntax, semantics, and pragmatics. Curiously, the enduring focus on the syntax and semantics in computational models has led to a subsequent neglect of pragmatics, an arguably essential pillar in meaning-making. Consequently, this impedes the capacity to appropriately understand and contextualize legal concepts.

---

[109] This is reminiscent of the argument raised in Kiel Brennan-Marquez and Stephen Henderson's article on concept of role-reversibility integral to the legal system. See Kiel Brennan-Marquez and Stephen Henderson, *Artificial Intelligence and Role-Reversible Judgment*, 109 J. Crim. L. & Criminology 137 (2019).

[110] Legislation as Code Report, *supra* 82 at 5.

[111] *Id.* at 83.

[112] The discussion by Sarah Lawsky furthers the support for form as a ripe area of formalization. *See* Sarah Lawsky, *Form as Formalization*, Ohio State Tech. L. J. (forthcoming 2020), available at: https://papers.ssrn.com/sol3/papers.cfm?abstract_id=3587576.

To recall, pragmatics is the field of linguistics that reflects on intention using tools of implicature and inference. Implicature, in linguistics, is defined as entailment, logically valid conclusions drawn between sentences.[113] Its counterpart, inference, is more complex. This is where discrepancies may exist, as what is being implied may differ from what is inferred. In accordance with Grice's Cooperative Principle,[114] the divergence between intended implicature and inference suggests non-conventional meaning. In effect, this supports the possibility of multiple interpretations on the basis of variations in context.

Consider the phrase: "There is an elephant in the tree." Semantics is helpful, to the extent, that it could raise what may be a prototype example of an elephant. As elephants are not typically found in trees, this is immediately a sign that this sentence may have a different meaning. Could this be a metaphorical idiom (i.e. elephant in the room) or perhaps there is some implicit understanding that the elephant in question is a paper elephant?  To recall, pragmatics raises the issue of reference. Consider the following sentences: "Jane is speaking with Joanne. She is a legal scholar."[115] The referent of "she" is not clear.  Without context, semantics alone cannot usefully provide information as to the meaning of these sentences.

There are parallels to the shortcomings of semantics revealed in propositional logic. Systems that use propositional logic, similar to Rules as Code structures, reflect the limitations presented in semantics. This is because propositional logic can enable the validation of some statements but cannot in itself establish the truth of all statements. So, why must there be consideration for pragmatics in machine-readable legislation?

Joseph A. Grundfest and A.C. Pritchard discuss the "technology of ambiguity" as a legislative strategy for compromise.[116] Their article reaffirms the notion of intentional, conscious, ambiguity.

As opposed to ambiguity as a 'bug,' Grundfest and Pritchard argue that it is feature of legislative drafting. That is, ambiguity in the drafting process is intended to work in tandem with the judiciary's interpretative methods. Ambiguity then works to ensure that the casuistic approach, characteristic of common law systems, is upheld.

Contrary to the rhetoric on clarity and precision, ambiguity is revered as an inherent property of statutory construction. While this is not necessarily a novel argument, Grundfest and Pritchard reassert the interoperability of the legal system; that legal documents are not independent artifacts and instead belong to a broader ecosystem. The aforementioned issues of pragmatics in natural language are integrated into the fabric of law and legal text and powered by literary tools of metaphor and analogy that outline context.

Interestingly, code is not quite as transparent or reducible as assumed. Mark C. Marino argues that code, like other systems of signification, cannot be removed from context. Code is not the result of mathematical certainty but "of collected cultural knowledge and convention (cultures of code and coding languages), haste and insight, inspirations and observations, evolutions and adaptations, rhetoric and reasons, paradigms of language, breakthroughs in approach, and failures to conceptualize."[117] While code appears to be 'solving' the woes of imprecision and lack of clarity in legal drafting, the use of code is, in fact, capturing meaning from a different paradigm. Rather, code is "frequently recontextualized" and meaning is "contingent upon and subject to the rhetorical triad of the speaker, audience (both human and machine), and message."[118] It follows that code is not a context-independent form of writing.  The questions become whether there could be a pragmatics of code, and if so, how could code effectively communicate legal concepts?

Marino articulates the "need to learn to read code critically."[119] Having understood the complexities and pitfalls of natural language, there is now a rising demand to understand the ways code acquires meaning and how shifting contexts shape and reshape this meaning. Currently, few scholars have addressed code beyond its operative capacity. This mirrors the focus on syntax and semantics as primary drivers of using code for legal drafting. Yet, learning how meaning is signified in code enables a deeper analysis of how the relationships, contexts, and requirements of law may be rightfully represented. From the science of (natural) language arises the science of code.

---

[117] Mark C. Marino, *Critical Code Studies* 8 (2020).

[118] *Id.* at 4.

[119] *Id.* at 5.

Increasingly, there has been emerging literature on the application of network analysis and graph theory to account for legal complexity. In a recent article on the growth of the law, representations of legislative materials were modelled using methods from network science and natural language processing.[120] The authors argue that quantifying law in a static manner fails to represent the diverse relationships and the interconnectivity of rules. They suggest that statutory materials should instead be represented using multidimensional, time-evolving document networks. As legal documents are interlinked, networks better reflect the dynamics of their language and the "deliberate design decisions made."[121] Moreover, it enables "circumvent[ing] some of the ambiguity problems that natural language-based approaches inherently face."[122] Most fascinating is the authors' capacity to isolate, through graph clustering techniques, legal topics that have fostered the most "complex bodies of legal rules."[123] This enabled a deeper understanding of the evolution of legal concepts and specific points of inflection where their perceptions have shifted.[124]

What is particularly striking about this paper is the introduction of quantitative approaches that stress content representation as opposed to structural miming. This model considers importantly context that shapes legal documents. How then could machine-readability be reconciled with graphical representation of legal documents? Statutory and legislative materials necessarily are situated at the heart of the legal ecosystem. That is, legislative documents provide the foundation on which other legal documents could gather concepts. This suggests that as opposed to an

---

emphasis on semantic translation to machine-readable legislation, a consideration of the role of legislation from an information extraction perspective may be a promising alternative.

## Conclusion

In analyzing the 'coming-of-age' of machine-readability, it becomes strikingly clear that, even with current advancements, there remains a gap around its role vis-à-vis 'human-readable' legislation.[125] The complexity of translating legislation from natural language to code stems from a persistent conceptualization of legal documents as independent entities. Taken in tandem with the practical ability of humans to write, code, understand, and judge, the notion that a perfect encoding of a rule might exist seems far-fetched. However, what is not far-fetched is encoding rules in a way that improves the ability of lawyers, judges, and other legal professionals to create antifragile systems and processes for disseminating legal knowledge.[126] That is, instead of focusing on creating the *perfect* encoding of a legal rule, what systems or processes could be created in order to more consistently extract meaning from statutes?

Legal information must be understood at a systemic level; to factor the interaction of legal documents with one another across a temporally sensitive frame. Therefore, legal texts should be perceived as objects with code as the semiotic vessel. How these objects interact, how references are made, and how their histories interrelate must be accounted for. It appears then that a dual-pronged method of semiotic analysis coupled with pragmatics contribute to a more fruitful engagement of legal knowledge representation. As opposed to applying an arithmetic lens in the name of clarity and precision, language design for machine-readability requires a multi-layered approach that extends beyond syntactic structure and ensures temporal management and formal ontological reference. Without these considerations, machine-readable legislation could only remain in the realm of a computable iteration.

---

[125] Building on this idea further, Douglas Hofstadter examines how difficult the concept of machine translation is in the broader context of general translation in his article in The Atlantic, "The Shallowness of Google Translate" https://www.theatlantic.com/technology/archive/2018/01/the-shallowness-of-google-translate/551570/.

[126] In Nassim Nicholas Taleb's book, *Antifragile: Things That Gain from Disorder*, he describes the ways antifragile systems are those that are neither robust (accounting for all possible circumstances) or resilience (withstanding all possible circumstances), but rather continually find ways to iterate such that cumulative knowledge can be created and used. *See* Nassim Nicholas Taleb, *Antifragile: Things That Gain from Disorder* (2012).

*Megan is the incoming CodeX Residential Fellow at Stanford Law School. She is also a PhD in Law candidate and Lecturer at Sciences Po, studying the relationship between law, language, and code. As well, she is the Managing Editor of the MIT Computational Law Report and has previously held Visiting positions at the University of Cambridge and Harvard Law School.*

*Bryan draws from a diversity of experience in academic and commercial contexts. As a Fellow in the MIT Connection Science research group under the direction of Professor Alex "Sandy" Pentland, Bryan works as Editor in Chief of the MIT Computational Law Report. As a Legal Engineer, Bryan works at Brighthive to create innovative data sharing products. Prior to that, Bryan was an early-stage employee at RiskGenius an exited InsurTech startup and was one of the inaugural fellows for the ABA Center for Innovation. This portfolio of efforts has earned him recognition by law.com as 1 of 18 Millennials Changing the Face of Legal Tech, as well as opportunities to speak at conferences throughout North America and Europe and teach courses on Computational Law and Revolutionary Ventures.*